\documentclass[11pt]{article}

\usepackage[final]{acl}

\usepackage{times}
\usepackage{latexsym}

\usepackage[T1]{fontenc}
\usepackage[utf8]{inputenc}

\usepackage{microtype}

\usepackage{inconsolata}

\usepackage{tikz}
\usetikzlibrary{arrows.meta, positioning, fit, backgrounds, calc}
\usepackage{xcolor}

\definecolor{clrGray}{RGB}{220,218,210}
\definecolor{clrTeal}{RGB}{29,158,117}
\definecolor{clrTealLight}{RGB}{225,245,238}
\definecolor{clrBlue}{RGB}{24,95,165}
\definecolor{clrBlueLight}{RGB}{230,241,251}
\definecolor{clrPurple}{RGB}{83,74,183}
\definecolor{clrPurpleLight}{RGB}{238,237,254}
\definecolor{clrCoral}{RGB}{216,90,48}
\definecolor{clrCoralLight}{RGB}{250,236,231}
\definecolor{clrAmber}{RGB}{186,117,23}
\definecolor{clrAmberLight}{RGB}{250,238,218}
\definecolor{clrGreen}{RGB}{59,109,17}
\definecolor{clrGreenLight}{RGB}{234,243,222}
\definecolor{clrGrayLight}{RGB}{241,239,232}

\usepackage{graphicx}
\usepackage{amsmath}
\usepackage{amssymb}
\usepackage{latexsym}
\usepackage{microtype}
\usepackage{booktabs}
\usepackage{multirow}
\usepackage{graphicx}
\usepackage{xcolor}
\usepackage{url}
\usepackage{hyperref}
\usepackage{enumitem}
\usepackage{xspace}
\usepackage{todonotes}
\usepackage{svg}
\usepackage{algorithm}
\usepackage{algorithmicx}
\usepackage{pifont}
\usepackage{orcidlink} %Added Orcid Id

\title{When Tokenizers Fail: Byte-Level Chunking for Zero-Shot Transfer to Low-Resource Languages}

\author{
  \textbf{Sanjeev Kumar}$^{1,2}$\,\orcidlink{0000-0002-4343-9334}\quad
  \textbf{Atsuki Yamaguchi}$^{2}$\,\orcidlink{0000-0001-8327-7598}\quad
  \textbf{Nikolaos Aletras}$^{2}$\,\orcidlink{0000-0003-4285-1965} \\[0.6em]
  $^{1}$Department of Computer Science and Engineering, IIT Bombay, India \\
  $^{2}$School of Computer Science, University of Sheffield, United Kingdom \\[0.5em]
  {\texttt{sanjeev@cse.iitb.ac.in}, \texttt{\{ayamaguchi1, n.aletras\}@sheffield.ac.uk}}
}
\DeclareUnicodeCharacter{2581}{\texttt{\char"2581}}
\begin{document}
\maketitle
\begin{abstract}
Subword tokenization hinders low-resource language processing by imposing frequency patterns from dominant languages onto script-sharing variants.
Byte-level models bypass this issue by processing raw UTF-8 characters, yet they create a granularity mismatch for word-level tasks in non-Latin scripts.
Hierarchical byte-level architectures address this mismatch by grouping bytes into word-aligned chunks.
However, these architectures require massive training data and suffer from representational misalignment when paired with frozen subword-based language models.
In this paper, we propose an adapted hierarchical network framework that bridges this modality gap without extensive training.
Our method initializes byte embeddings directly from the subword representations of a frozen base model.
We apply a chunk alignment loss to project dynamically grouped byte chunks toward precomputed subword targets, and interleave lightweight part-of-speech (POS) supervision to guide boundary detection.
Experiments across six languages demonstrate that our tokenizer-free approach improves performance for word-level morphological tasks, yielding up to a 13.3\% improvement on POS tagging.\footnote{Code: \href{https://github.com/snjev310/ByteChunk}{\texttt{github.com/snjev310/ByteChunk}}.}
\end{abstract}

\section{Introduction} \label{sec:intro}
%%%%%
% Background
%%%%%
Natural language processing (NLP) for low-resource languages is fundamentally constrained by tokenization. Modern multilingual models, such as Qwen3~\citep{yang2025qwen3technicalreport} and Gemma3~\citep{gemmateam2025gemma3technicalreport}, rely on subword tokenizers trained on large corpora, typically using algorithms such as Byte Pair Encoding (BPE)~\cite{sennrich-etal-2016-neural} or unigram language models (LMs)~\citep{kudo-2018-subword}.
While these tokenizers capture meaningful morphological structures for well-represented languages, they often perform poorly for low-resource languages due to limited vocabulary coverage and segmentation mismatches~\cite{bostrom-durrett-2020-byte,xue-etal-2022-byt5,clark-etal-2022-canine}.

In this work, we focus explicitly on Indic languages. The Indic language family serves as an ideal testbed because these languages feature high morphological complexity and variable script overlap, making them susceptible to tokenizer overfragmentation.
For instance, Indic languages such as Bhojpuri and Magahi share the Devanagari script with resource-rich Hindi, causing tokenizers to apply Hindi-derived subwords.
Table~\ref{tab:fragmentation} provides quantitative evidence of this tokenization mismatch.
The Qwen3 tokenizer splits 97.7\% of Bhojpuri words into multiple subword tokens, with an average of 4.43 tokens per word.
Given modern subword tokenizers do not generate out-of-vocabulary tokens due to byte-fallback mechanisms, the tokenizer forces representations into available subwords, masking segmentation failures beneath valid, yet linguistically incorrect, token sequences.
Consequently, models must operate on fragmented and linguistically misaligned input units, degrading performance on downstream tasks.

% \begin{table*}[t]
% \centering
% \small
% \setlength{\tabcolsep}{3pt}
% \begin{tabular}{lrrrrrrrr}
% \toprule
% \multirow{2}{*}{\textbf{Language}} &
% \multirow{2}{*}{\textbf{Words}} &
% \multicolumn{3}{c}{\textbf{Qwen Subword}} &
% \multicolumn{3}{c}{\textbf{H-Net Chunk}} \\
% \cmidrule(lr){3-5}\cmidrule(lr){6-8}
% & &
% \textbf{Subwords} & \textbf{\% Split} & \textbf{SW/W} &
% \textbf{Chunks} & \textbf{C/W} & \textbf{1:1 Align\%} \\
% \midrule
% Hindi    & 281,057 & 1,155,744 & 93.7\% & 4.11 & 348,202 & 1.24 & 92.5\% \\
% Bhojpuri &   6,665 &    29,491 & 97.7\% & 4.43 &   7,785 & 1.17 & 91.0\% \\
% Marathi  &   2,997 &    12,917 & 81.6\% & 4.31 &   4,267 & 1.42 & 82.5\% \\
% Magahi   &   7,702 &    37,740 & 97.2\% & 4.90 &   9,042 & 1.17 & 87.8\% \\
% Sanskrit &   1,843 &    10,542 & 98.9\% & 5.72 &   3,621 & 1.96 & 87.3\% \\
% Urdu     &  14,806 &    44,714 & 95.4\% & 3.02 &  15,666 & 1.06 & 93.2\% \\
% \bottomrule
% \end{tabular}
% \caption{Tokenization and chunking statistics across six Indic languages.
% \textbf{Qwen Subword}: Subwords = total subword tokens produced;
% \% Split = words fragmented into $\geq$2 subword tokens;
% SW/W = subword tokens per word. UNK rate is 0\% for all languages
% --- the tokenizer silently applies incorrect morphological patterns
% rather than signaling vocabulary failure.
% \textbf{H-Net Chunk}: Chunks = total chunks produced without
% any word segmentation supervision; C/W = chunks per word
% (ideally $\approx$1.0); 1:1 Align\% = percentage of chunks
% covering exactly one CoNLL-U word (82--93\% across all languages).}
% \label{tab:fragmentation}
% \end{table*}

\begin{table*}[t]
\centering
\small
\renewcommand{\arraystretch}{0.8}
\setlength{\aboverulesep}{1.3pt}
\setlength{\belowrulesep}{1.3pt}
\setlength{\tabcolsep}{4pt}
\resizebox{0.94\textwidth}{!}{
\begin{tabular}{lrrrrrrrrrrr}
\toprule
& & \multicolumn{3}{c}{\textbf{Qwen3 Subword}} 
& \multicolumn{3}{c}{\textbf{Gemma3 Subword}}
& \multicolumn{2}{c}{\textbf{H-Net (Qwen3)}}
& \multicolumn{2}{c}{\textbf{H-Net (Gemma3)}} \\
\cmidrule(lr){3-5}\cmidrule(lr){6-8}\cmidrule(lr){9-10}\cmidrule(lr){11-12}
\textbf{Language} & \textbf{Words} 
& \textbf{\#SW} & \textbf{SW/W} & \textbf{Split (\%)}
& \textbf{\#SW} & \textbf{SW/W} & \textbf{Split (\%)}
& \textbf{C/W} & \textbf{1:1 (\%)}
& \textbf{C/W} & \textbf{1:1 (\%)} \\
\midrule
Hindi    & 281,057 & 1,155,744 & 4.11 & 93.7 & 445,511 & 1.59 & 42.1 & 1.24 & 92.5 & 0.98 & 90.4 \\
\midrule
Bhojpuri &   6,665 &    29,491 & 4.43 & 97.7 &  11,708 & 1.76 & 52.9 & 1.17 & 91.0 & 0.95 & 79.6 \\
Marathi  &   2,997 &    12,917 & 4.31 & 81.6 &   5,354 & 1.79 & 56.0 & 1.42 & 82.5 & 0.94 & 89.0 \\
Magahi   &   7,702 &    37,740 & 4.90 & 97.2 &  13,401 & 1.74 & 52.5 & 1.17 & 87.8 & 0.97 & 91.5 \\
Sanskrit &   1,843 &    10,542 & 5.72 & 98.9 &   4,533 & 2.46 & 75.9 & 1.96 & 87.3 & 1.00 & 99.6 \\
Urdu     &  14,806 &    44,714 & 3.02 & 95.4 &  28,403 & 1.92 & 60.0 & 1.06 & 93.2 & 0.59 & 0.2 \\
\bottomrule
\end{tabular}
}
\caption{Tokenizer fragmentation and H-Net chunk alignment statistics across six Indic languages. \textbf{\#SW} stands for the total number of subwords. \textbf{SW/W} and \textbf{C/W} denote the subword-to-word and chunk-to-word ratios, respectively. \textbf{Split (\%)} indicates the percentage of words fragmented into two or more subword tokens, while \textbf{1:1 (\%)} represents the percentage of chunks covering exactly one word.}
%Qwen fragments Devanagari languages at 4.11--5.72 subwords/word;
%Gemma achieves near word-level tokenization at 1.59--2.46 subwords/word.
%H-Net (Gemma backbone) achieves near-perfect chunk-word alignment for Devanagari languages (C/W $\approx$ 1.0) but fails for Urdu ($^\dagger$C/W=0.59, 1:1\%=0.2\%) due to script mismatch between Nastaliq training and Devanagari pretraining data.
\label{tab:fragmentation}
\end{table*}

%%%%%
% Current solution
%%%%%
A natural alternative is to eliminate the dependency on subword tokenization entirely.
Byte-level models operate directly on raw UTF-8 representations, avoiding segmentation errors and providing robustness across languages~\cite{xue-etal-2022-byt5,clark-etal-2022-canine}.
However, operating at the byte level introduces a granularity mismatch for word-level tasks.
Non-Latin scripts such as Devanagari encode each character using multiple UTF-8 bytes, which dilutes the word-level signal across multiple positions. Hierarchical byte-level architectures, such as H-Net~\cite{hwang2026dynamic} and Bolmo~\cite{minixhofer2026bolmobyteifyinggenerationlanguage}, address this issue by dynamically grouping bytes into word-aligned chunks.

%%%%%
% Research gap
%%%%%
However, existing hierarchical paradigms are not suitable for adapting models to low-resource languages.
They rely on (pre-)training with massive token budgets, offering no feasible path for data-constrained scenarios.
Furthermore, naively combining a frozen subword-based backbone with randomly initialized byte-level components creates a representational mismatch. The resulting byte-level chunk embeddings lack explicit alignment with the subword semantic space of the backbone, which prevents effective cross-lingual transfer.

%%%%%
% What we have done to solve the issue
%%%%%
In this paper, we introduce an adapted H-Net framework, a novel hybrid approach that bridges this modality gap without requiring large-scale training.
First, our method initializes byte embeddings directly from the subword embeddings of the backbone.
Second, it applies a chunk alignment loss to pull byte-level chunk embeddings toward pre-computed subword alignment targets.
Unlike unguided byte encoding that produces incompatible representations, our alignment mechanism ensures that the frozen backbone can effectively interpret the newly formed chunks.
Finally, we interleave lightweight part-of-speech (POS) supervision during training to guide boundary detection.
This process allows our adapted model to bypass tokenization bottlenecks while retaining the robust linguistic knowledge of the pretrained backbone.

We evaluate our approach on POS tagging, named entity recognition (NER), and sentiment analysis across five Indic languages using two model families at three parameter scales (1.7B, 4B, and 12B). Our contributions are:
\begin{enumerate}[leftmargin=*,itemsep=2pt]
% Method + Broader impact
\item A novel adapted hierarchical framework that bridges byte-level inputs and frozen subword LMs without extensive training, addressing the bottleneck of tokenizer overfragmentation in low-resource languages. (\S\ref{sec:approach})

% Experiments
\item The proposed architecture consistently improves zero-shot cross-lingual transfer across languages. It outperforms a subword baseline on POS tagging by up to 13.3\%, achieving highly competitive target-language word-level morphological processing capabilities. (\S\ref{sec:results})

% Analysis - Ablation (Why does our approach work? 
\item  Our analysis shows that backbone-initialized embeddings prevent training collapse under limited data, task-guided pretraining improves chunk-level representations by up to 16.4 points, and tokenizer-aware alignment targets are essential for SentencePiece backbones. (\S\ref{sec:analysis})
\end{enumerate}

\section{Related Work} 
\label{sec:related}

\paragraph{Byte-level LMs.}

Byte-level models eliminate multilingual NLP vocabulary bottlenecks by operating directly on raw UTF-8 bytes \citep{xue-etal-2022-byt5}. Previous work utilizes hierarchical architectures to manage long sequences \citep{NEURIPS2023_f8f78f80, pagnoni-etal-2025-byte}, group adjacent representations \citep{hwang2026dynamic}, or align distinct modalities \citep{minixhofer2026bolmobyteifyinggenerationlanguage}. However, these methodologies are strictly constrained to high-resource settings and demand massive training corpora that range from 43B to 4T bytes. Furthermore, current frameworks focus primarily on Latin scripts, require tokenizer-derived boundaries, or rely on large-scale end-to-end pretraining of all components, rendering them impractical for low-resource target languages. Related work on vocabulary adaptation \cite{downey-etal-2023-embedding} transfers embedding structure across subword vocabularies,
\citet{feher-etal-2025-retrofitting} retrofit LLMs with dynamic tokenization, and \citet{remy2024transtokenization} adapt subword vocabularies for low-resource NLP.

Unlike these approaches, which all operate in the subword space, our framework processes raw bytes and learns word-aligned chunks without any tokenizer.
Specifically, the proposed method targets low-resource Indic languages by adapting off-the-shelf, subword-based LMs. We replace the original embeddings and LM heads with H-Net byte-level modules. We then train only these newly introduced components on a minimal corpus of 500K sentences ($\sim$ 1B bytes), leaving the backbone frozen.

\paragraph{Tokenization \& low-resource languages.}
Prior work identifies structural limitations of subword tokenization in low-resource settings~\cite{ahia-etal-2023-languages,NEURIPS2023_74bb24dc}.
\citet{rust-etal-2021-good} show that BPE tokenizers overfragment text in low-resource languages, and \citet{adelani-etal-2022-masakhaner} report analogous segmentation failures for African languages.
\citet{foroutan-etal-2026-parity} further establish that standard BPE vocabulary selection inherently disadvantages underrepresented languages under data constraints.
For Indic languages, \citet{pattnayak2025tokenization} observe that BPE tokenizers degrade zero-shot cross-lingual NER, and \citet{kumar-etal-2024-part} demonstrate that POS tagging suffers directly from tokenizer overfragmentation. \citet{brahma-etal-2026-multilingual} attribute these vulnerabilities to script diversity, complex morphology, and restricted pretraining coverage.
Building on these observations, we quantify this severe tokenizer fragmentation across target Indic languages in Table~\ref{tab:fragmentation}, motivating the design of our byte-level framework.

\paragraph{Cross-lingual transfer.}
Cross-lingual transfer relies on shared vocabulary and tokenizer coverage. \citet{pires-etal-2019-multilingual} show that shared subword vocabularies facilitate transfer in multilingual BERT, while \citet{wu-dredze-2020-languages} find that vocabulary overlap correlates with transfer quality.
Similar observations emerge across models, where lexical overlap and tokenizer coverage substantially affect cross-lingual representation alignment and downstream transfer performance \citep{conneau-etal-2020-unsupervised,devlin-etal-2019-bert,rust-etal-2021-good,yamaguchi-etal-2024-empirical,yamaguchi-etal-2026-effectively}.
Consequently, for low-resource languages lacking tokenizer coverage, traditional transfer mechanisms degrade rapidly, isolating them from the benefits of pretrained models.
To resolve this dependency, the proposed framework bypasses the requirement for subword overlap entirely.
By aligning byte-level chunks directly with the semantic space of a frozen subword backbone, the proposed framework enables robust cross-lingual transfer without relying on shared tokenization.\looseness=-1

\section{Methodology}
\begin{figure*}[t]
    \centering
    \includegraphics[width=0.95\textwidth,height=7cm]{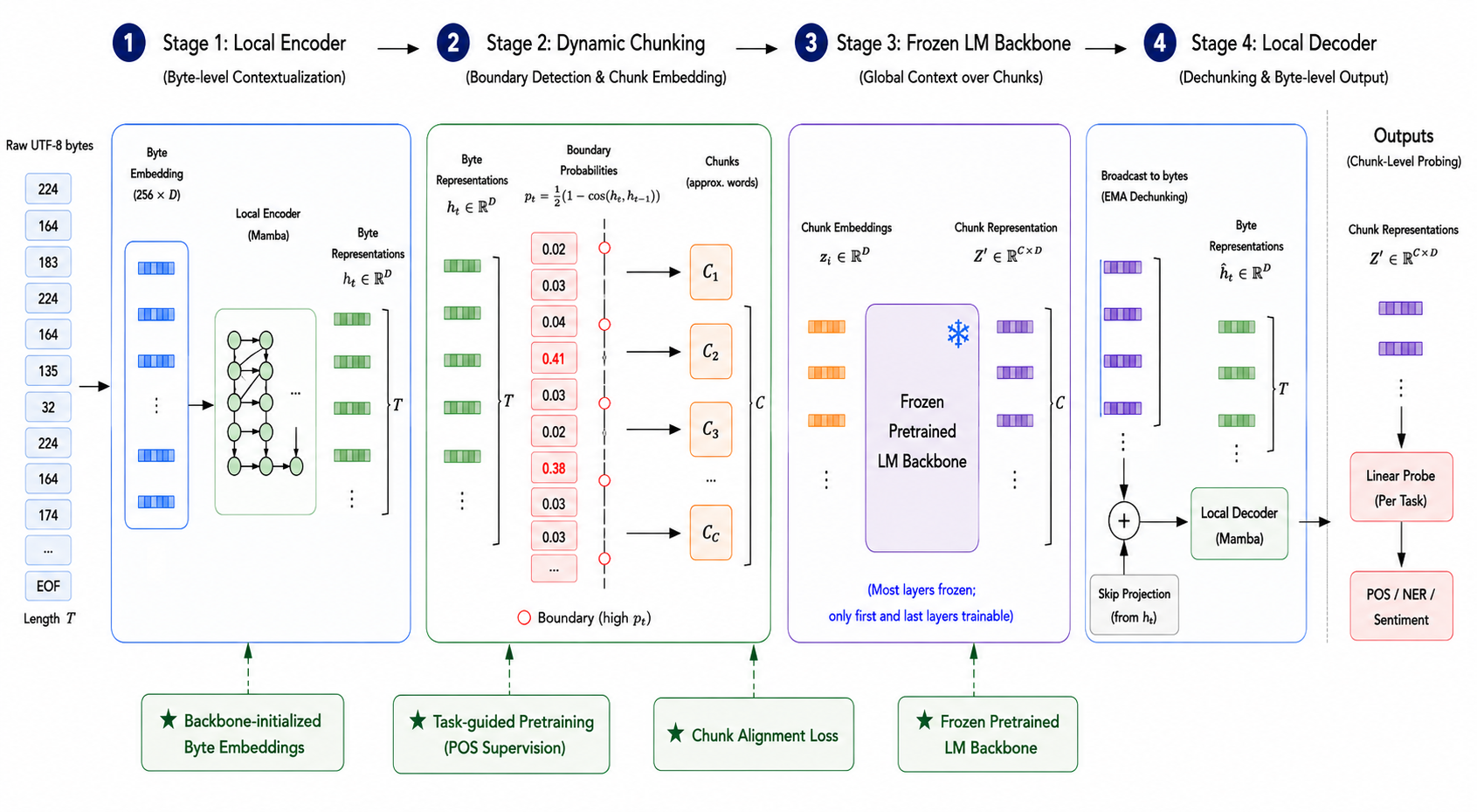}
    \caption{
    Overview of the proposed H-Net architecture for low-resource languages.}
    % Raw UTF-8 bytes are encoded using Mamba layers, dynamically grouped into variable-length chunks based on cosine-similarity boundary detection, processed globally by a pretrained backbone, and then dechunked back to byte-level representations for downstream tasks.}
    \label{fig:hnet_pipeline}
\end{figure*}

\subsection{Preliminaries: H-Net}
\label{sec:hnet_prelim}
H-Net~\citep{hwang2026dynamic} is a hierarchical byte-level architecture that processes raw UTF-8 bytes directly, bypassing subword tokenization.
Given a byte sequence, H-Net operates in four stages.

\paragraph{Stage 1: Local encoder.}
First, the model projects each input byte $x$ to a $D$-dimensional vector via a learned embedding table $\mathbf{E} \in \mathbb{R}^{256 \times D}$: $\mathbf{h}^{(0)} = \mathbf{E}[\mathbf{x}] \in \mathbb{R}^{T \times D}$,
where $T$ denotes the total sequence length. A local encoder comprising $L$ sequential Mamba~\citep{gu2024mamba} layers then contextualizes these embeddings using local neighborhood information. The representation at the $l$-th layer is updated via a residual connection:
\begin{equation*}
  \mathbf{h}^{(l)} = \mathbf{h}^{(l-1)} + \text{Mamba}(\mathbf{h}^{(l-1)}) \in \mathbb{R}^{T \times D},
\end{equation*}
where $l \in \{1, \dots, L\}$. The final output $\mathbf{h} = \mathbf{h}^{(L)}$ provides the contextualized byte representations necessary to capture the short-range dependencies required for boundary detection.

\paragraph{Stage 2: Dynamic chunking.}
Next, the module partitions bytes into variable-length chunks. At each position $t$, the system computes a boundary probability based on the cosine similarity between adjacent byte representations:
\begin{equation*}
  p_t = \frac{1}{2}\!\left(1 - \cos(\mathbf{h}_t,\,
  \mathbf{h}_{t-1})\right).
\end{equation*}
A high $p_t$ value indicates a probable word boundary, whereas a low $p_t$ value implies a within-word continuation. The model assigns hard boundaries where $p_t \geq \tau$. For each chunk, the module selects the byte representation at the boundary position as the initial chunk embedding.

A smoothing module subsequently refines this representation by applying self-attention across all byte representations contained within the chunk:
\begin{equation*}
  \mathbf{z}_i = \text{Smooth}(\mathbf{h}_{b_i})
  \in \mathbb{R}^{D},
  \label{eq:chunk_emb}
\end{equation*}
where $b_i$ denotes the boundary position of chunk $i$.

\paragraph{Stage 3: LM backbone.}
The sequence of $C$ chunk embeddings $\mathbf{Z} = (\mathbf{z}_1, \ldots, \mathbf{z}_C) \in \mathbb{R}^{C \times D}$ serves as input to a global LM backbone:
\begin{equation}
  \mathbf{Z}' = \text{Backbone}(\mathbf{Z}) \in \mathbb{R}^{C \times D}.
  \label{eq:lm}
\end{equation}
Because the backbone processes chunk-level representations rather than subword tokens, it avoids tokenizer-induced overfragmentation.

\paragraph{Stage 4: Local decoder.} \label{eq:dechunk}
Finally, the Exponential Moving Average (EMA) DeChunk operation~\citep{hwang2026dynamic} distributes chunk representations back to byte positions using exponentially decaying weights determined by the distance from chunk boundaries:
\begin{multline}
  \hat{\mathbf{h}} = \text{Mamba}\big(
    \text{EMADeChunk}(\mathbf{Z}') \\
    + \text{SkipProj}(\mathbf{h})\big) 
  \in \mathbb{R}^{T \times D}
  \label{eq:decoder}
\end{multline}
These final representations integrate local byte-level information with global chunk-level context.

\paragraph{Pretraining objectives.}
The architecture optimizes two primary losses. It employs standard next-byte prediction via a linear byte-level LM head:
\begin{equation*}
  \mathcal{L}_{\text{AR}} = -\sum_{t=1}^{T-1} \log p(x_{t+1} \mid \hat{\mathbf{h}}_t).
\end{equation*}

To maintain optimal chunk sizes, a boundary ratio loss regularizes the mean boundary probability toward a target rate $\rho^*$, which approximately corresponds to one boundary per word:
\begin{equation}
  \mathcal{L}_{\text{ratio}} = \left(\bar{p} - \rho^*\right)^2,
  \label{eq:ratio_loss}
\end{equation}
where $\bar{p} = \frac{1}{T}\sum_t p_t$ defines the mean boundary probability.
The complete H-Net pretraining loss combines these terms:
\begin{equation*}
  \mathcal{L}_{\text{H-Net}} = \mathcal{L}_{\text{AR}}
  + \lambda_{\text{ratio}}\,\mathcal{L}_{\text{ratio}},
\end{equation*}
where $\lambda_{\text{ratio}}$ is a hyperparameter controlling the boundary regularization penalty.

\subsection{Adapting H-Net for low-resource languages}
\label{sec:approach}
We adapt H-Net by reusing a frozen, subword-based pretrained LM backbone rather than pretraining the entire architecture from scratch. Although this strategy provides a computationally efficient foundation, applying it to low-resource languages poses three challenges: slow convergence caused by random byte embedding initialization, a lack of explicit word-level supervision for boundary detection, and a representational gap between byte-level chunk embeddings and the semantic space of the backbone.
We introduce three targeted modifications to resolve each issue.

\paragraph{Backbone-initialized byte embeddings.}
We replace random initialization by averaging the backbone subword embeddings that contain each specific byte value:
\begin{equation}
  \mathbf{E}[b] = \frac{1}{|\mathcal{T}_b|}
  \sum_{v \in \mathcal{T}_b} \mathbf{W}_v,
  \label{eq:init}
\end{equation}
where $\mathcal{T}_b$ is the set of subword tokens containing byte $b$, and $\mathbf{W}_v$ is the backbone embedding for token $v$.
This strategy places the byte representations within the pretrained semantic space before training, accelerating convergence under limited data.\looseness=-1

\paragraph{Chunk alignment loss.}
To align byte-level chunks with the backbone embedding space, we introduce a loss function that pulls each chunk embedding toward a byte alignment target, defined as the average of the corresponding backbone subword embeddings:
\begin{equation}
  \mathcal{L}_{\text{align}} = \frac{1}{C}\sum_{i=1}^{C}
  \left(1 - \cos(\mathbf{z}_i,\, \bar{\mathbf{t}}_i)\right),
  \label{eq:chunk}
\end{equation}
where $\bar{\mathbf{t}}_i$ is the target for chunk $i$. This ensures the frozen backbone ``interprets'' the new chunk embeddings without requiring extensive fine-tuning, thereby facilitating effective transfer.

\paragraph{Task-guided pretraining.}
The standard boundary ratio loss (Eq. \ref{eq:ratio_loss}) often fails to learn linguistic boundaries in low-resource settings due to insufficient supervision signals.
To mitigate this, we interleave POS supervision on a target language. Every $k$ steps (with $k=10$ in this study), the model trains a linear POS head jointly with the encoder:\looseness=-1
\begin{equation}
  \mathcal{L}_{\text{POS}} = -\sum_{t \in \mathcal{V}}
  \log p(\text{tag}_t \mid \hat{\mathbf{h}}_t),
  \label{eq:task-guided-pt}
\end{equation}
where $\mathcal{V}$ defines the valid byte positions (excluding padding and spaces).
The two objectives use separate batches and forward/backward passes. Autoregressive steps process FineWeb2 Hindi text with $\mathcal{L} = \mathcal{L}_{\text{AR}} +\lambda_{\text{ratio}} \mathcal{L}_{\text{ratio}} +\lambda_{\text{align}} \mathcal{L}_{\text{align}}$, while POS steps process Hindi CoNLL-U sentences with $\mathcal{L} = \lambda_{\text{POS}}\mathcal{L}_{\text{POS}}$ only.

We discard this head after pretraining. The POS data acts as a structural signal for learning span representations~\citep{tenney-etal-2019-bert,hewitt-manning-2019-structural}.

\paragraph{Full pretraining objective.}
The complete pretraining loss combines our novel contributions ($\mathcal{L}_{\text{POS}}$, $\mathcal{L}_{\text{align}}$) with the standard H-Net methodology ($\mathcal{L}_{\text{AR}}, \mathcal{L}_{\text{ratio}}$):
\begin{equation*}
\begin{split}
  \mathcal{L}_{\text{total}} &= \mathcal{L}_{\text{AR}}
  + \lambda_{\text{ratio}}\,\mathcal{L}_{\text{ratio}} \\
  &\quad + 
  \lambda_{\text{POS}}\,\mathcal{L}_{\text{POS}}
  + \lambda_{\text{align}}\,\mathcal{L}_{\text{align}}
  ,
\end{split}
\end{equation*}
where $\lambda_{\text{POS}}$ and $\lambda_{\text{align}}$ are hyperparameters controlling the POS supervision and alignment terms.

\section{Experimental Setup} \label{sec:exp_setup}
\paragraph{Target languages.}
We evaluate the proposed framework on five low-resource Indic languages: Bhojpuri, Marathi, Magahi, Sanskrit, and Urdu.
We select Hindi as the source adaptation language because it offers sufficient training data while sharing linguistic roots, structural characteristics, or the Devanagari script with the target variants.
This typological proximity allows Hindi to function as an effective anchor for cross-lingual transfer.

\paragraph{Models.}
We utilize pretrained base model variants from  Qwen3~\cite{yang2025qwen3technicalreport} and Gemma3~\cite{gemmateam2025gemma3technicalreport}, with two scale variants from each model family: Qwen3-1.7B, Qwen3-4B, Gemma3-4B, and Gemma3-12B.
These model families provide a clear contrast in their multilingual pretraining corpora, as Gemma3 incorporates broader multilingual data than Qwen3. Qwen3 is pretrained on 119 languages, whereas Gemma3 is pretrained on 140 languages. 
Neither model reports explicit pretraining coverage for Bhojpuri, Magahi, and Sanskrit. Marathi is likely covered in both families, as it uses the Devanagari script and is closely related to Hindi, which is well represented in both pretraining corpora. Urdu is likely covered because it uses the Nastaliq script, which is also used by Arabic and Persian.

\paragraph{Training.}
We use 500K Hindi sentences from the FineWeb2 \texttt{hin\_Deva} subset (1.02B bytes)~\citep{DBLP:journals/corr/abs-2506-20920} for adaptation.
During training, we unfreeze the first and the last transformer layers of the backbone model to optimize computational efficiency while ensuring effective adaptation~\citep{remy2024transtokenization,yamaguchi2025adapting}. For task-guided pretraining (Eq. \ref{eq:task-guided-pt}), we use POS data from the Universal Dependencies (UD) Hindi treebank~\citep{nivre-etal-2020-universal} (\texttt{hi-hdtb}, 13K sentences).
Following adaptation, we train a two-layer MLP probe on frozen chunk representations for downstream evaluation. Full hyperparameter details are reported in Appendix~\ref{sec:training_details}.

\paragraph{Baselines.}
We compare the proposed approach (H-Net) against two baseline configurations:
\begin{itemize}
    \item \textbf{Subword} (Sub): The frozen pretrained backbone acts as a feature extractor. For word-level tasks, we train a two-layer MLP probe on the last-layer hidden states. For the sentence-level task, we train the probe on the mean-pooled last-layer hidden states over all non-padding tokens.

    \item \textbf{Continued pretraining} (CPT): We continually pretrain the backbone on the identical 500K Hindi sentences used for the proposed method. We unfreeze the first and the last layers to match the setup as the proposed method. This baseline isolates the contribution of the byte-level architecture by controlling for the effect of additional Hindi exposure.\looseness=-1
\end{itemize}

\begin{table*}[t]
\centering
\small
\setlength{\tabcolsep}{3pt}
\begin{tabular}{lllcccccccccccc}
\toprule
& \multirow{2}{*}{\textbf{Language}} &
\multirow{2}{*}{\textbf{Script}} &
\multicolumn{3}{c}{\textbf{Qwen3-1.7B}} &
\multicolumn{3}{c}{\textbf{Qwen3-4B}} &
\multicolumn{3}{c}{\textbf{Gemma3-4B}} &
\multicolumn{3}{c}{\textbf{Gemma3-12B}} \\
\cmidrule(lr){4-6}\cmidrule(lr){7-9}\cmidrule(lr){10-12}\cmidrule(lr){13-15}
& & & Sub & CPT & H-Net & Sub & CPT & H-Net & Sub & CPT & H-Net & Sub & CPT & H-Net \\
\midrule
\addlinespace[0.3em]
\multirow{7}{*}{\large \ding{172}}& \multicolumn{14}{c}{\textbf{Part-of-Speech Tagging (Accuracy \%)}} \\
\cmidrule{2-15}
& Bhojpuri & Devanagari & 43.8 & 48.1 &\textbf{54.4} & 47.9 & 49.3 & \textbf{57.9} & \textbf{59.2} & 56.9 & 56.9 & 55.2 & 56.4 &\textbf{56.8} \\
& Marathi  & Devanagari & 39.1 & 35.3 &\textbf{46.3} & 44.0 & 36.8 & \textbf{46.0} & 55.6 & 44.5 & \textbf{59.2} & 53.9 & 41.7 &\textbf{58.4} \\
& Magahi   & Devanagari & 41.0 & 40.9 &\textbf{53.1} & 45.3 & 43.2 & \textbf{46.5} & \textbf{55.3} & 51.9 & 53.1 & 49.8 & 50.4 &\textbf{52.3} \\
& Sanskrit & Devanagari & 21.6 & 30.2 &\textbf{30.6} & 25.7 & \textbf{35.1} & 33.2 & 21.9 & \textbf{32.8} & 30.6 & 30.6 & 31.1 &\textbf{32.7} \\
& Urdu     & Nastaliq   & 38.8 & 48.1 &\textbf{52.1} & 48.6 & 51.0 & \textbf{55.6} & 46.1 & \textbf{52.4} & 52.1 & \textbf{70.2} & 65.6 &53.1 \\
\cmidrule{2-15}
& Average  &            & 36.9 & 40.5 &\textbf{47.3} & 42.3 & 43.1 & \textbf{47.8} & 47.6 & 47.7 & \textbf{50.4} & \textbf{51.9} & 49.0 &50.7 \\
\midrule
\addlinespace[0.3em]
\multirow{5}{*}{\large \ding{173}} & \multicolumn{14}{c}{\textbf{Named Entity Recognition (Entity-level F1 \%)}} \\
\cmidrule{2-15}
& Urdu     & Nastaliq   & \textbf{12.7} & 5.2 & 12.1 & 2.1  & 4.0  & \textbf{14.4} & 8.1  & 3.8  & \textbf{9.4}  & 5.7  & 4.9 & \textbf{12.8} \\
& Marathi  & Devanagari & \textbf{14.8} & 14.2 & 14.3 & 13.0 & 13.0 & \textbf{13.3} & 22.4 & 16.6 & \textbf{27.7} & 29.1 & 21.6 & \textbf{32.5} \\
& Sanskrit & Devanagari & 9.3  & 10.4 & \textbf{19.9} & 10.5 & 10.2 & \textbf{15.4} & 15.1 & 10.7 & \textbf{19.3} & 12.2 & \textbf{14.6} & 14.5 \\
\cmidrule{2-15}
& Average  &            & 12.3 & 9.9 & \textbf{15.4} & 8.5  & 9.1  & \textbf{14.4} & 15.2 & 10.4 & \textbf{18.8} & 15.7 & 13.7 & \textbf{19.9} \\
\midrule
\addlinespace[0.3em]
\multirow{4}{*}{\large \ding{174}} & \multicolumn{14}{c}{\textbf{Sentiment Analysis (Accuracy \%)}} \\
\cmidrule{2-15}
& Marathi  & Devanagari & \textbf{72.7} & 53.6 & 63.3 & \textbf{54.4} & 54.3 & \textbf{54.4} & \textbf{91.8} & 85.3 & 78.0 & \textbf{90.0} & 85.5 & 75.0 \\
& Urdu     & Nastaliq   & \textbf{49.3} & \textbf{49.3} & \textbf{49.3} & 49.3 & 49.5 & \textbf{51.0} & \textbf{69.3} & 50.0 & 65.4 & \textbf{84.0} & 79.8 & 68.0 \\
\cmidrule{2-15}
& Average  &            & \textbf{61.0} & 51.5 & 56.3 & 51.9 & 51.9 & \textbf{52.7} & \textbf{80.6} & 67.7 & 71.7 & \textbf{87.0} & 82.7 & 71.5 \\
\bottomrule
\end{tabular}
\caption{Zero-shot performance across three downstream tasks.
Bold indicates the best result among Sub, CPT, and H-Net per task and model family.}
\label{tab:zs_all_tasks}
\end{table*}

\paragraph{Evaluation tasks, metrics, and protocols.}
We primarily evaluate the models on two word-level sequence labeling tasks: \textbf{POS tagging} using UD treebanks~\citep{nivre-etal-2020-universal}, and \textbf{NER} using WikiANN~\citep{rahimi-etal-2019-massively}. These tasks directly assess whether the byte-level chunks of the adapted architecture produce linguistically meaningful word-level representations.
As a supplementary evaluation, we include \textbf{sentiment analysis} using IndicSentiment~\citep{doddapaneni-etal-2023-towards}, a sentence-level semantic task, to examine whether the byte-level representations generalize beyond morphological structure to broader semantic understanding. For POS tagging, we report token-level accuracy. For NER, we report entity-level F1. For sentiment analysis, we report accuracy.

For the proposed framework, we evaluate at the chunk level: each contextualized chunk embedding $\mathbf{z}'_i \in \mathbf{Z}'$ (Eq.~\ref{eq:lm}) approximates one word and receives the label of the word that contributes the majority of its bytes. Chunks covering only spaces or punctuation are excluded ($\text{ignore\_index}{=}{-}100$). For subword baselines, we evaluate only the first subword token per word \cite{pires-etal-2019-multilingual, wolf-etal-2020-transformers, wu-dredze-2020-languages}.

We conduct all evaluations in a zero-shot setting. We apply a probe trained on Hindi representations directly to the target languages without any target-language supervision.
Further details on dataset statistics, language morphology, and script characteristics are provided in Appendix~\ref{sec:lang_complexity}.

\section{Results} \label{sec:results}

\paragraph{POS tagging.} Our adapted H-Net models consistently outperform the corresponding subword baselines on POS tagging (Table~\ref{tab:zs_all_tasks}~\ding{172}).
Specifically, H-Net (Qwen3-1.7B) surpasses the subword baseline in all five low-resource languages with an average gain of 10.4 points, and H-Net (Qwen3-4B) achieves an average gain of 5.5 points.
Similarly, H-Net (Gemma3-4B) outperforms the subword baseline on three out of five languages (+2.8 points average), with the maximal gain observed on Sanskrit (+8.7 points).
In contrast, CPT yields negligible gains across models providing an average increase of only 0.8 points for Qwen3-4B and 0.1 points for Gemma3-4B.
This confirms that additional Hindi exposure alone cannot resolve the tokenizer bottleneck.

The advantage of H-Net peaks where tokenizer overfragmentation is most severe.
Magahi provides a clear example of this dynamic. The Qwen3 tokenizer splits Magahi words into an average of 4.90 tokens (Table~\ref{tab:fragmentation}), causing severe degradation for the subword models, whereas H-Net (Qwen3-1.7B) achieves a 12.1 point performance gain.

Beyond fragmentation, results on Urdu suggest that byte-level transfer in H-Net can generalize beyond shared byte sequences. Despite the use of the Perso-Arabic Nastaliq script, H-Net (Qwen3-1.7B) achieves 52.1\% zero-shot accuracy (+13.3 points). We hypothesize that linguistic relatedness plays a role in morphological transfer even when byte patterns differ.

\paragraph{NER.} 
This structural advantage of our adapted H-Net models extends to NER.
Table~\ref{tab:zs_all_tasks}~\ding{173} shows that H-Net broadly outperforms the subword baseline (Table~\ref{tab:zs_all_tasks}).
The gains average 3.1 points for Qwen3-1.7B and 3.6 points for Gemma3-4B, reaching a maximum improvement on Sanskrit (10.6 points, Qwen3-1.7B), where tokenization is most fragmented (5.72 tokens/word; Table \ref{tab:fragmentation}).
Conversely, CPT often underperforms both the proposed method and the subword baseline across this task. This failure reinforces our finding in POS tagging: simply exposing a model to more source-language data without hierarchical byte-level boundary adaptation degrades the cross-lingual representation of entity-level semantics. \looseness=-1

\paragraph{Sentiment analysis.}
Unlike the word-level tasks, subword baselines dominate sentence-level sentiment classification (Table~\ref{tab:zs_all_tasks}~\ding{174}).
Larger subword models (Gemma3-4B at 80.6\%, Gemma3-12B at 87.0\%) outperform all H-Net variants.
In addition, CPT underperforms the subword baseline across different backbones.
This outcome aligns with the finding that semantic tasks benefit from pretraining scale rather than Hindi-specific adaptation~\cite{JMLR:v21:20-074, kaplan2020scalinglawsneurallanguage,xue-etal-2021-mt5}. H-Net, adapted on only 500K Hindi sentences, has substantially less exposure to sentiment-bearing vocabulary than Qwen3 or Gemma3, which are trained on trillions of tokens across hundreds of languages. All Qwen3 models collapse to near-random performance on Urdu ($\sim$49\%), indicating that sentiment-bearing lexical cues do not transfer reliably across divergent scripts such as Devanagari and Nastaliq in a zero-shot setting. 

\paragraph{Takeaways.}
H-Net achieves peak performance for word-level tasks (POS and NER), successfully bypassing tokenizer fragmentation to capture syntactic and entity-level structures. 
However, it does not outperform the subword baseline in sentiment analysis.
Furthermore, CPT consistently underperforms H-Net across tasks, confirming that the observed gains arise directly from the H-Net architecture rather than additional Hindi training exposure.
Overall, these results support our hypothesis (\S\ref{sec:intro}): hierarchical byte-level models benefit tasks that depend on accurate morphological structure, such as POS tagging and NER, where word-boundary quality directly affects label prediction. In contrast, sentiment analysis relies more on opinion-bearing vocabulary and semantics derived from large-scale pretraining corpora. In-language Hindi evaluation is provided in \S\ref{app:hindi}.\looseness=-1

\section{Analysis} \label{sec:analysis}

This section verifies the specific contribution of each architectural adaptation introduced in \S\ref{sec:approach}.
Due to resource constraints, we use only Qwen3-1.7B for our analysis.

\begin{figure}[t]
\centering
\includegraphics[width=\linewidth]{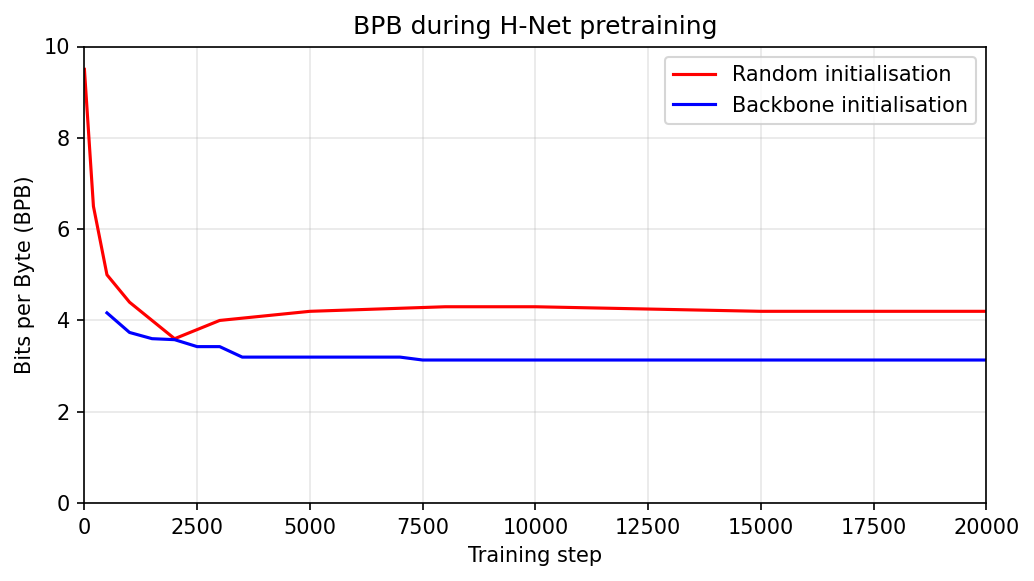}
\caption{BPB during H-Net pretraining under random (\textcolor{red}{red}) vs backbone (\textcolor{blue}{blue}) byte embedding initialization.} 
% Random initialization starts at BPB $\approx 9.5$, drops to $\sim$3.6, then rises and plateaus at $4.2$. Backbone initialization starts at BPB $= 4.17$ and converges smoothly to $3.13$ within 7,500 steps ($-$25\% relative), remaining stable thereafter.}
\label{fig:init_comparison}
\end{figure}

\begin{figure*}
    \centering
    \includegraphics[width=0.95\linewidth,height=5cm]{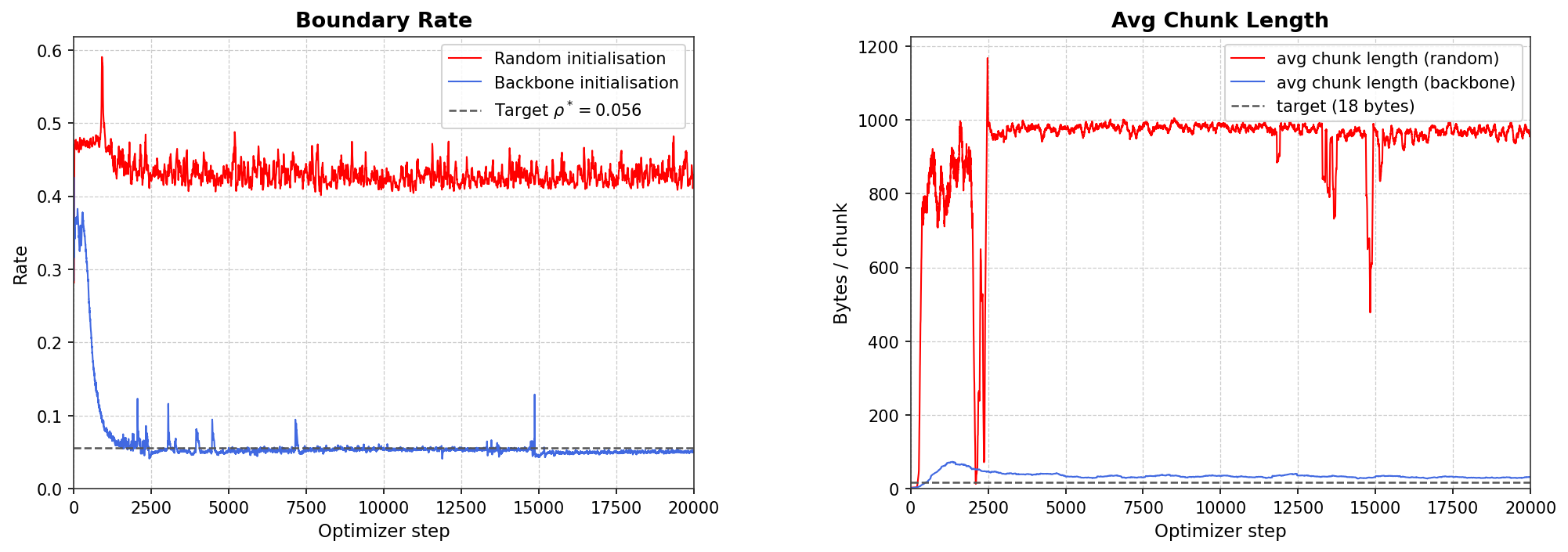}
    \caption{Training dynamics with random (\textcolor{red}{red}) vs backbone (\textcolor{blue}{blue}) byte embedding initialization. \textbf{Left}: average boundary ratio over 20,000 steps oscillates at 0.43--0.49. \textbf{Right}: average chunk length.}
    \label{fig:random_init}
\end{figure*}

\paragraph{Backbone-initialized byte embeddings.}

We evaluate the necessity of initializing byte embeddings directly from the subword representations of the backbone (Eq. \ref{eq:init}).
Standard random initialization introduces an immediate semantic mismatch with the pretrained backbone, which severely hinders training dynamics (Figure~\ref{fig:init_comparison}).
Specifically, the bits-per-byte (BPB) metric for  the random initialization model begins at approximately 9.5, drops sharply to 3.6, and subsequently rises to plateau at 4.2, showing no further improvement over 20K steps.
In contrast, our backbone initialization mitigates this mismatch from the outset; it starts at a BPB of 4.17 and converges smoothly to 3.13 within 7.5K steps, achieving a 25\% relative reduction.

The autoregressive loss (i.e., BPB) converges normally in both cases (Figure~\ref{fig:init_comparison}), confirming that the model memorizes Hindi byte sequences regardless of initialization.
However, the routing module fails to learn word-level boundaries under random initialization: the boundary ratio fluctuates between 0.43 and 0.49 throughout training (Figure~\ref{fig:random_init}, left).
This constitutes a substantial deviation from the target of $\rho^{*} = 0.056$, which corresponds to the expected frequency of one boundary per word.
%(target: $\rho^{*} = 0.056$), and average chunk length remains 200--1200 bytes (target: $\sim$18 bytes/word) (Figure~\ref{fig:random_init}).
Consequently, the average chunk length remains between 200 and 1200 bytes, well above the target of 18 bytes per word (Figure~\ref{fig:random_init}, right).
Backbone initialization instead places byte embeddings in the correct semantic neighborhood from the first training step, the boundary ratio drops from 0.50 at initialization to $\approx$0.05 (target: $\rho^{*} = 0.056$), providing a meaningful gradient signal to the alignment loss immediately.\looseness=-1

\paragraph{Task-guided pretraining.}
\begin{table}[t]
\renewcommand{\arraystretch}{0.8}
\centering
\small
\begin{tabular}{lcc}
\toprule
\textbf{Pretraining} & \textbf{Byte-level} & \textbf{Chunk-level} \\
\midrule
Without POS & 32.5 & 43.1 \\
With POS    & 32.8 & 59.5 \\
\bottomrule
\end{tabular}
\caption{Hindi POS accuracy (\%) with and without task-guided pretraining (Qwen3-1.7B backbone).}
\label{tab:ablation_pos}
\end{table}
We next evaluate the impact of task-guided pretraining on the structural quality of chunk boundaries.
Table \ref{tab:ablation_pos} demonstrates that introducing explicit POS supervision improves chunk-level accuracy by 16.4 points (from 43.1\% to 59.5\%) while maintaining raw byte prediction capabilities (a 0.3 point increase in byte-level accuracy). This gain highlights the limitations of relying solely on the default boundary ratio loss.
As defined in \S\ref{sec:hnet_prelim}, the default objective (Eq. \ref{eq:ratio_loss}) targets a global average boundary rate ($\rho^{*} = 0.056$, corresponding to approximately one boundary per word). While the network successfully achieves this global statistical target, it frequently misaligns with true linguistic boundaries at the local level due to insufficient structural signals.
Interleaving POS supervision resolves this misalignment by forcing the encoder to match gold standard Hindi word boundaries. Furthermore, this structural improvement generalizes across tasks. The framework trained with POS supervision outperforms subword baselines on NER (Table~\ref{tab:zs_all_tasks}), proving that linguistically structured boundaries transfer effectively even in the absence of task-specific morphological supervision.\looseness=-1

\paragraph{Chunk alignment loss.} \label{sec:design_alignment}
\begin{table}[t]
\centering
\small
\renewcommand{\arraystretch}{0.8}
\setlength{\aboverulesep}{1.3pt}
\setlength{\belowrulesep}{1.3pt}
\setlength{\tabcolsep}{4pt}
\begin{tabular}{lcccc}
\toprule
\multirow{2}{*}{\textbf{Language}} &
\multicolumn{2}{c}{\textbf{Full model}} &
\multicolumn{2}{c}{\textbf{w/o align loss}} \\
\cmidrule(lr){2-3}\cmidrule(lr){4-5}
& C/W & 1:1\% & C/W & 1:1\% \\
\midrule
Hindi    & 1.24 & 92.5 & 0.05 & 5.5 \\
Bhojpuri & 1.17 & 91.0 & 0.06 & 3.4 \\
Marathi  & 1.42 & 82.5 & 0.13 & 0.0 \\
Magahi   & 1.17 & 87.8 & 0.07 & 0.0 \\
Sanskrit & 1.96 & 87.3 & 0.13 & 0.4 \\
Urdu     & 1.06 & 93.2 & 0.07 & 27.5 \\
\bottomrule
\end{tabular}
\caption{Chunk-word alignment statistics for the full H-Net model vs without chunk alignment loss ($\lambda_{\text{align}} = 0$). }
% C/W = chunks per word; 1:1\% = percentage of words aligned to exactly one chunk. $-$ = results pending.}
\label{tab:ablation_align}
\end{table}
Finally, we analyze the necessity of the chunk alignment loss (Eq. \ref{eq:chunk}) in bridging the representational gap between the newly introduced byte-level modules and the frozen backbone.
Table~\ref{tab:ablation_align} shows that without the alignment loss, C/W collapses to a range of 0.06 to 0.13 across all six languages.
This indicates that each chunk spans between 8 and 17 words on average, heavily deviating from the target ratio of approximately 1.0. 
Furthermore, the 1:1\% metric drops to near zero, reflecting a complete failure of word-level boundary detection.
These results demonstrate that without explicit alignment, chunk embeddings occupy a completely different region of the embedding space compared to the subword tokens the backbone processed during pretraining. Consequently, the frozen backbone produces degraded contextualized representations.

The root cause of this boundary detection failure is the geometry of the byte alignment targets determined by the backbone tokenizer.
The Qwen3 tokenizer assigns the space character a dedicated token with a distinct embedding, creating a space byte target geometrically separate from Devanagari byte targets (cosine similarity of 0.40).
In contrast, the Gemma3 tokenizer absorbs spaces into word-initial tokens (e.g., \texttt{\_word}). Consequently, the isolated space byte lacks a dedicated embedding and receives a near-zero target vector (norm of 0.007).
This vector yields a cosine similarity of 0.487 with Devanagari byte targets, eliminating the boundary signal ($p_{\text{max}} = 0.16$).
While the 0.087 similarity difference between Qwen3 and Gemma3 appears small, a value of 0.487 approaches the 0.5 random baseline for these embeddings, and masks the boundary entirely.
This lack of contrast prevents the routing module from distinguishing genuine space-induced word boundaries from standard within-word byte continuations. To resolve this limitation, the proposed method computes the space target as the centroid of all word-initial token embeddings. This intervention restores geometric contrast between space and text bytes (cosine similarity of 0.032) and recovers word-level chunking.

\section{Conclusion}
We presented an adapted H-Net framework that addresses the tokenizer overfragmentation bottleneck for low-resource Indic languages.
By combining backbone-initialized byte embeddings, interleaved POS supervision, and a tokenizer-aware chunk alignment loss, the proposed method produces word-aligned chunk representations from only 500K Hindi sentences.
Experiments across five languages show that the adapted architecture improves performance on word-level morphological tasks, yielding average gains of 10.4 points on POS tagging and 7.8 points on NER.
This framework provides a computationally efficient mechanism to extend the structural knowledge of high-resource LMs to extremely low-resource languages, entirely bypassing the dependency on shared tokenization vocabularies.

\section*{Limitations}
\paragraph{Pretraining language.}
In our current setup, we only consider Indic low-resource languages as our family of interest. H-Net is pretrained exclusively on Hindi, a high-resource Devanagari language. Transfer to target languages relies on script sharing (Bhojpuri, Magahi, Marathi, Sanskrit) or linguistic relatedness (Urdu). Languages that share neither Hindi's script nor its morphological structure are unlikely to benefit from this approach without additional pretraining data.

\paragraph{Script coverage.} Our evaluation covers only two scripts: Devanagari and Nastaliq. H-Net's boundary detection exploits Devanagari-specific byte patterns (matra structure), and its effectiveness for other non-Latin scripts, such as Tamil, Chinese, or Bengali, remains to be validated.

\paragraph{Task scope.} Our evaluation covers POS tagging, NER, and sentiment analysis tasks. Performance on tasks requiring deeper semantic understanding, such as machine translation or natural language inference, remains unexplored. Evaluation on machine translation using FLORES-200~\citep{goyal-etal-2022-flores} or IndicMT Eval~\citep{sai-b-etal-2023-indicmt}, potentially combined with related-source mixing~\citep{kumar-etal-2026-srcmix}, is left for future work.

\paragraph{Efficiency.} We do not report wall-clock inference time. H-Net processes raw bytes via a local Mamba encoder, which incurs additional computational overhead compared to subword tokenization. However, dynamic chunking reduces the sequence length fed to the backbone by ${\sim}18{\times}$, partially compensating for this overhead. A systematic efficiency comparison is left for future work.

\paragraph{Compute asymmetry.} H-Net and the CPT baseline use the same training data but different compute budgets. H-Net processes 1.02B byte tokens while CPT processes approximately 7.5M subword tokens. A fully compute-matched comparison remains an open question.

\section*{Acknowledgements}
We thank the anonymous reviewers for their insightful and constructive comments.
This work was conducted while SK was a visiting student at the University of Sheffield. SK thanks Prof. Preethi Jyothi (IIT Bombay) for supporting the research visit. SK gratefully acknowledges support from the Amazon--IITB AI/ML initiative towards conference expenses, the IBM AI Horizon Networks--IIT Bombay initiative for financial support towards the visit to the University of Sheffield, and the TCS Research Foundation for Ph.D. fellowship support for conducting research on extremely low-resource Indian languages.
AY is supported by the Engineering and Physical Sciences Research Council (EPSRC) [grant number EP/W524360/1] and the Japan Student Services Organization (JASSO) Student Exchange Support Program (Graduate Scholarship for Degree Seeking Students).

\bibliography{custom,latex/anthology-1,latex/anthology-2}

\begin{thebibliography}{43}
\providecommand{\natexlab}[1]{#1}

\bibitem[{Adelani et~al.(2022)Adelani, Neubig, Ruder, Rijhwani, Beukman, Palen-Michel, Lignos, Alabi, Muhammad, Nabende, Dione, Bukula, Mabuya, Dossou, Sibanda, Buzaaba, Mukiibi, Kalipe, Mbaye, Taylor, Kabore, Emezue, Aremu, Ogayo, Gitau, Munkoh-Buabeng, Memdjokam~Koagne, Tapo, Macucwa, Marivate, Mboning, Gwadabe, Adewumi, Ahia, Nakatumba-Nabende, Mokono, Ezeani, Chukwuneke, Adeyemi, Hacheme, Abdulmumin, Ogundepo, Yousuf, Moteu~Ngoli, and Klakow}]{adelani-etal-2022-masakhaner}
David~Ifeoluwa Adelani, Graham Neubig, Sebastian Ruder, Shruti Rijhwani, Michael Beukman, Chester Palen-Michel, Constantine Lignos, Jesujoba~O. Alabi, Shamsuddeen~H. Muhammad, Peter Nabende, Cheikh M.~Bamba Dione, Andiswa Bukula, Rooweither Mabuya, Bonaventure F.~P. Dossou, Blessing Sibanda, Happy Buzaaba, Jonathan Mukiibi, Godson Kalipe, Derguene Mbaye, and 26 others. 2022.
\newblock \href {https://doi.org/10.18653/v1/2022.emnlp-main.298} {{M}asakha{NER} 2.0: {A}frica-centric transfer learning for named entity recognition}.
\newblock In \emph{Proceedings of the 2022 Conference on Empirical Methods in Natural Language Processing}, pages 4488--4508, Abu Dhabi, United Arab Emirates. Association for Computational Linguistics.

\bibitem[{Ahia et~al.(2023)Ahia, Kumar, Gonen, Kasai, Mortensen, Smith, and Tsvetkov}]{ahia-etal-2023-languages}
Orevaoghene Ahia, Sachin Kumar, Hila Gonen, Jungo Kasai, David Mortensen, Noah Smith, and Yulia Tsvetkov. 2023.
\newblock \href {https://doi.org/10.18653/v1/2023.emnlp-main.614} {Do all languages cost the same? tokenization in the era of commercial language models}.
\newblock In \emph{Proceedings of the 2023 Conference on Empirical Methods in Natural Language Processing}, pages 9904--9923, Singapore. Association for Computational Linguistics.

\bibitem[{Bostrom and Durrett(2020)}]{bostrom-durrett-2020-byte}
Kaj Bostrom and Greg Durrett. 2020.
\newblock \href {https://doi.org/10.18653/v1/2020.findings-emnlp.414} {Byte pair encoding is suboptimal for language model pretraining}.
\newblock In \emph{Findings of the Association for Computational Linguistics: EMNLP 2020}, pages 4617--4624, Online. Association for Computational Linguistics.

\bibitem[{Brahma et~al.(2026)Brahma, Karthika, Verma, Naidu, Saluja, Desarkar, and Ramakrishnan}]{brahma-etal-2026-multilingual}
Maharaj Brahma, N~J Karthika, Rajat Verma, Nagasai~Saketh Naidu, Rohit Saluja, Maunendra~Sankar Desarkar, and Ganesh Ramakrishnan. 2026.
\newblock \href {https://doi.org/10.18653/v1/2026.findings-acl.1632} {Multilingual tokenization through the lens of {I}ndian languages: Challenges and insights}.
\newblock In \emph{Findings of the {A}ssociation for {C}omputational {L}inguistics: {ACL} 2026}, pages 32614--32632, San Diego, California, United States. Association for Computational Linguistics.

\bibitem[{Clark et~al.(2022)Clark, Garrette, Turc, and Wieting}]{clark-etal-2022-canine}
Jonathan~H. Clark, Dan Garrette, Iulia Turc, and John Wieting. 2022.
\newblock \href {https://doi.org/10.1162/tacl_a_00448} {Canine: Pre-training an efficient tokenization-free encoder for language representation}.
\newblock \emph{Transactions of the Association for Computational Linguistics}, 10:73--91.

\bibitem[{Conneau et~al.(2020)Conneau, Khandelwal, Goyal, Chaudhary, Wenzek, Guzm{\'a}n, Grave, Ott, Zettlemoyer, and Stoyanov}]{conneau-etal-2020-unsupervised}
Alexis Conneau, Kartikay Khandelwal, Naman Goyal, Vishrav Chaudhary, Guillaume Wenzek, Francisco Guzm{\'a}n, Edouard Grave, Myle Ott, Luke Zettlemoyer, and Veselin Stoyanov. 2020.
\newblock \href {https://doi.org/10.18653/v1/2020.acl-main.747} {Unsupervised cross-lingual representation learning at scale}.
\newblock In \emph{Proceedings of the 58th Annual Meeting of the Association for Computational Linguistics}, pages 8440--8451, Online. Association for Computational Linguistics.

\bibitem[{Devlin et~al.(2019)Devlin, Chang, Lee, and Toutanova}]{devlin-etal-2019-bert}
Jacob Devlin, Ming-Wei Chang, Kenton Lee, and Kristina Toutanova. 2019.
\newblock \href {https://doi.org/10.18653/v1/N19-1423} {{BERT}: Pre-training of deep bidirectional transformers for language understanding}.
\newblock In \emph{Proceedings of the 2019 Conference of the North {A}merican Chapter of the Association for Computational Linguistics: Human Language Technologies, Volume 1 (Long and Short Papers)}, pages 4171--4186, Minneapolis, Minnesota. Association for Computational Linguistics.

\bibitem[{Doddapaneni et~al.(2023)Doddapaneni, Aralikatte, Ramesh, Goyal, Khapra, Kunchukuttan, and Kumar}]{doddapaneni-etal-2023-towards}
Sumanth Doddapaneni, Rahul Aralikatte, Gowtham Ramesh, Shreya Goyal, Mitesh~M. Khapra, Anoop Kunchukuttan, and Pratyush Kumar. 2023.
\newblock \href {https://doi.org/10.18653/v1/2023.acl-long.693} {Towards leaving no {I}ndic language behind: Building monolingual corpora, benchmark and models for {I}ndic languages}.
\newblock In \emph{Proceedings of the 61st Annual Meeting of the Association for Computational Linguistics (Volume 1: Long Papers)}, pages 12402--12426, Toronto, Canada. Association for Computational Linguistics.

\bibitem[{Downey et~al.(2023)Downey, Blevins, Goldfine, and Steinert-Threlkeld}]{downey-etal-2023-embedding}
C.M. Downey, Terra Blevins, Nora Goldfine, and Shane Steinert-Threlkeld. 2023.
\newblock \href {https://doi.org/10.18653/v1/2023.mrl-1.20} {Embedding structure matters: Comparing methods to adapt multilingual vocabularies to new languages}.
\newblock In \emph{Proceedings of the 3rd Workshop on Multi-lingual Representation Learning (MRL)}, pages 268--281, Singapore. Association for Computational Linguistics.

\bibitem[{Feher et~al.(2025)Feher, Vuli{\'c}, and Minixhofer}]{feher-etal-2025-retrofitting}
Darius Feher, Ivan Vuli{\'c}, and Benjamin Minixhofer. 2025.
\newblock \href {https://doi.org/10.18653/v1/2025.acl-long.1444} {Retrofitting large language models with dynamic tokenization}.
\newblock In \emph{Proceedings of the 63rd Annual Meeting of the Association for Computational Linguistics (Volume 1: Long Papers)}, pages 29866--29883, Vienna, Austria. Association for Computational Linguistics.

\bibitem[{Foroutan et~al.(2026)Foroutan, Meister, Paul, Niklaus, Ahmadi, Bosselut, and Sennrich}]{foroutan-etal-2026-parity}
Negar Foroutan, Clara Meister, Debjit Paul, Joel Niklaus, Sina Ahmadi, Antoine Bosselut, and Rico Sennrich. 2026.
\newblock \href {https://doi.org/10.18653/v1/2026.acl-long.342} {Parity-aware byte-pair encoding: Improving cross-lingual fairness in tokenization}.
\newblock In \emph{Proceedings of the 64th Annual Meeting of the {A}ssociation for {C}omputational {L}inguistics (Volume 1: Long Papers)}, pages 7514--7538, San Diego, California, United States. Association for Computational Linguistics.

\bibitem[{{Gemma Team} et~al.(2025){Gemma Team}, Kamath, Ferret, Pathak, Vieillard, Merhej, Perrin, Matejovicova, Ramé, Rivière, Rouillard, Mesnard, Cideron, bastien Grill, Ramos, Yvinec, Casbon, Pot, Penchev, Liu, Visin, Kenealy, Beyer, Zhai, Tsitsulin, Busa-Fekete, Feng, Sachdeva, Coleman, Gao, Mustafa, Barr, Parisotto, Tian, Eyal, Cherry, Peter, Sinopalnikov, Bhupatiraju, Agarwal, Kazemi, Malkin, Kumar, Vilar, Brusilovsky, Luo, Steiner, Friesen, Sharma, Sharma, Gilady, Goedeckemeyer, Saade, Feng, Kolesnikov, Bendebury, Abdagic, Vadi, György, Pinto, Das, Bapna, Miech, Yang, Paterson, Shenoy, Chakrabarti, Piot, Wu, Shahriari, Petrini, Chen, Lan, Choquette-Choo, Carey, Brick, Deutsch, Eisenbud, Cattle, Cheng, Paparas, Sreepathihalli, Reid, Tran, Zelle, Noland, Huizenga, Kharitonov, Liu, Amirkhanyan, Cameron, Hashemi, Klimczak-Plucińska, Singh, Mehta, Lehri, Hazimeh, Ballantyne, Szpektor, Nardini, Pouget-Abadie, Chan, Stanton, Wieting, Lai, Orbay, Fernandez, Newlan, yeong Ji, Singh, Black, Yu, Hui,
  Vodrahalli, Greff, Qiu, Valentine, Coelho, Ritter, Hoffman, Watson, Chaturvedi, Moynihan, Ma, Babar, Noy, Byrd, Roy, Momchev, Chauhan, Sachdeva, Bunyan, Botarda, Caron, Rubenstein, Culliton, Schmid, Sessa, Xu, Stanczyk, Tafti, Shivanna, Wu, Pan, Rokni, Willoughby, Vallu, Mullins, Jerome, Smoot, Girgin, Iqbal, Reddy, Sheth, Põder, Bhatnagar, Panyam, Eiger, Zhang, Liu, Yacovone, Liechty, Kalra, Evci, Misra, Roseberry, Feinberg, Kolesnikov, Han, Kwon, Chen, Chow, Zhu, Wei, Egyed, Cotruta, Giang, Kirk, Rao, Black, Babar, Lo, Moreira, Martins, Sanseviero, Gonzalez, Gleicher, Warkentin, Mirrokni, Senter, Collins, Barral, Ghahramani, Hadsell, Matias, Sculley, Petrov, Fiedel, Shazeer, Vinyals, Dean, Hassabis, Kavukcuoglu, Farabet, Buchatskaya, Alayrac, Anil, Dmitry, Lepikhin, Borgeaud, Bachem, Joulin, Andreev, Hardin, Dadashi, and Hussenot}]{gemmateam2025gemma3technicalreport}
{Gemma Team}, Aishwarya Kamath, Johan Ferret, Shreya Pathak, Nino Vieillard, Ramona Merhej, Sarah Perrin, Tatiana Matejovicova, Alexandre Ramé, Morgane Rivière, Louis Rouillard, Thomas Mesnard, Geoffrey Cideron, Jean bastien Grill, Sabela Ramos, Edouard Yvinec, Michelle Casbon, Etienne Pot, Ivo Penchev, and 197 others. 2025.
\newblock \href {https://arxiv.org/abs/2503.19786} {Gemma 3 technical report}.
\newblock \emph{Preprint}, arXiv:2503.19786.

\bibitem[{Goyal et~al.(2022)Goyal, Gao, Chaudhary, Chen, Wenzek, Ju, Krishnan, Ranzato, Guzm{\'a}n, and Fan}]{goyal-etal-2022-flores}
Naman Goyal, Cynthia Gao, Vishrav Chaudhary, Peng-Jen Chen, Guillaume Wenzek, Da~Ju, Sanjana Krishnan, Marc{'}Aurelio Ranzato, Francisco Guzm{\'a}n, and Angela Fan. 2022.
\newblock \href {https://doi.org/10.1162/tacl_a_00474} {The {F}lores-101 evaluation benchmark for low-resource and multilingual machine translation}.
\newblock \emph{Transactions of the Association for Computational Linguistics}, 10:522--538.

\bibitem[{Gu and Dao(2024)}]{gu2024mamba}
Albert Gu and Tri Dao. 2024.
\newblock \href {https://openreview.net/forum?id=tEYskw1VY2} {Mamba: Linear-time sequence modeling with selective state spaces}.
\newblock In \emph{First Conference on Language Modeling}.

\bibitem[{Hewitt and Manning(2019)}]{hewitt-manning-2019-structural}
John Hewitt and Christopher~D. Manning. 2019.
\newblock \href {https://doi.org/10.18653/v1/N19-1419} {{A} structural probe for finding syntax in word representations}.
\newblock In \emph{Proceedings of the 2019 Conference of the North {A}merican Chapter of the Association for Computational Linguistics: Human Language Technologies, Volume 1 (Long and Short Papers)}, pages 4129--4138, Minneapolis, Minnesota. Association for Computational Linguistics.

\bibitem[{Hwang et~al.(2026)Hwang, Wang, and Gu}]{hwang2026dynamic}
Sukjun Hwang, Brandon Wang, and Albert Gu. 2026.
\newblock \href {https://openreview.net/forum?id=ZbfLR9NbNF} {Dynamic chunking for end-to-end hierarchical sequence modeling}.
\newblock In \emph{The Fourteenth International Conference on Learning Representations}.

\bibitem[{Kaplan et~al.(2020)Kaplan, McCandlish, Henighan, Brown, Chess, Child, Gray, Radford, Wu, and Amodei}]{kaplan2020scalinglawsneurallanguage}
Jared Kaplan, Sam McCandlish, Tom Henighan, Tom~B. Brown, Benjamin Chess, Rewon Child, Scott Gray, Alec Radford, Jeffrey Wu, and Dario Amodei. 2020.
\newblock \href {https://arxiv.org/abs/2001.08361} {Scaling laws for neural language models}.
\newblock \emph{Preprint}, arXiv:2001.08361.

\bibitem[{Kudo(2018)}]{kudo-2018-subword}
Taku Kudo. 2018.
\newblock \href {https://doi.org/10.18653/v1/P18-1007} {Subword regularization: Improving neural network translation models with multiple subword candidates}.
\newblock In \emph{Proceedings of the 56th Annual Meeting of the Association for Computational Linguistics (Volume 1: Long Papers)}, pages 66--75, Melbourne, Australia. Association for Computational Linguistics.

\bibitem[{Kumar et~al.(2024)Kumar, Jyothi, and Bhattacharyya}]{kumar-etal-2024-part}
Sanjeev Kumar, Preethi Jyothi, and Pushpak Bhattacharyya. 2024.
\newblock \href {https://doi.org/10.18653/v1/2024.findings-acl.857} {Part-of-speech tagging for extremely low-resource {I}ndian languages}.
\newblock In \emph{Findings of the Association for Computational Linguistics: ACL 2024}, pages 14422--14431, Bangkok, Thailand. Association for Computational Linguistics.

\bibitem[{Kumar et~al.(2026)Kumar, Jyothi, and Bhattacharyya}]{kumar-etal-2026-srcmix}
Sanjeev Kumar, Preethi Jyothi, and Pushpak Bhattacharyya. 2026.
\newblock \href {https://doi.org/10.18653/v1/2026.findings-eacl.332} {{S}rc{M}ix: Mixing of related source languages benefits extremely low-resource machine translation}.
\newblock In \emph{Findings of the {A}ssociation for {C}omputational {L}inguistics: {EACL} 2026}, pages 6306--6323, Rabat, Morocco. Association for Computational Linguistics.

\bibitem[{Minixhofer et~al.(2026)Minixhofer, Murray, Limisiewicz, Korhonen, Zettlemoyer, Smith, Ponti, Soldaini, and Hofmann}]{minixhofer2026bolmobyteifyinggenerationlanguage}
Benjamin Minixhofer, Tyler Murray, Tomasz Limisiewicz, Anna Korhonen, Luke Zettlemoyer, Noah~A. Smith, Edoardo~M. Ponti, Luca Soldaini, and Valentin Hofmann. 2026.
\newblock \href {https://arxiv.org/abs/2512.15586} {Bolmo: Byteifying the next generation of language models}.
\newblock \emph{Preprint}, arXiv:2512.15586.

\bibitem[{Nivre et~al.(2020)Nivre, de~Marneffe, Ginter, Haji{\v{c}}, Manning, Pyysalo, Schuster, Tyers, and Zeman}]{nivre-etal-2020-universal}
Joakim Nivre, Marie-Catherine de~Marneffe, Filip Ginter, Jan Haji{\v{c}}, Christopher~D. Manning, Sampo Pyysalo, Sebastian Schuster, Francis Tyers, and Daniel Zeman. 2020.
\newblock \href {https://aclanthology.org/2020.lrec-1.497/} {{U}niversal {D}ependencies v2: An evergrowing multilingual treebank collection}.
\newblock In \emph{Proceedings of the Twelfth Language Resources and Evaluation Conference}, pages 4034--4043, Marseille, France. European Language Resources Association.

\bibitem[{Pagnoni et~al.(2025)Pagnoni, Pasunuru, Rodriguez, Nguyen, Muller, Li, Zhou, Yu, Weston, Zettlemoyer, Ghosh, Lewis, Holtzman, and Iyer}]{pagnoni-etal-2025-byte}
Artidoro Pagnoni, Ramakanth Pasunuru, Pedro Rodriguez, John Nguyen, Benjamin Muller, Margaret Li, Chunting Zhou, Lili Yu, Jason~E Weston, Luke Zettlemoyer, Gargi Ghosh, Mike Lewis, Ari Holtzman, and Srini Iyer. 2025.
\newblock \href {https://doi.org/10.18653/v1/2025.acl-long.453} {Byte latent transformer: Patches scale better than tokens}.
\newblock In \emph{Proceedings of the 63rd Annual Meeting of the Association for Computational Linguistics (Volume 1: Long Papers)}, pages 9238--9258, Vienna, Austria. Association for Computational Linguistics.

\bibitem[{Pattnayak et~al.(2025)Pattnayak, Patel, and Agarwal}]{pattnayak2025tokenization}
Priyaranjan Pattnayak, Hitesh Patel, and Amit Agarwal. 2025.
\newblock \href {https://doi.org/10.1109/eIT64391.2025.11103625} {Tokenization matters: Improving zero-shot ner for indic languages}.
\newblock In \emph{2025 IEEE International Conference on Electro Information Technology (eIT)}, pages 456--462.

\bibitem[{Penedo et~al.(2025)Penedo, Kydl{\'\i}{\v{c}}ek, Sabol{\v{c}}ec, Messmer, Foroutan, Kargaran, Raffel, Jaggi, Werra, and Wolf}]{DBLP:journals/corr/abs-2506-20920}
Guilherme Penedo, Hynek Kydl{\'\i}{\v{c}}ek, Vinko Sabol{\v{c}}ec, Bettina Messmer, Negar Foroutan, Amir~Hossein Kargaran, Colin Raffel, Martin Jaggi, Leandro~Von Werra, and Thomas Wolf. 2025.
\newblock \href {https://openreview.net/forum?id=jnRBe6zatP} {{FineWeb2}: One pipeline to scale them all {\textemdash} adapting pre-training data processing to every language}.
\newblock In \emph{Second Conference on Language Modeling}.

\bibitem[{Petrov et~al.(2023)Petrov, La~Malfa, Torr, and Bibi}]{NEURIPS2023_74bb24dc}
Aleksandar Petrov, Emanuele La~Malfa, Philip Torr, and Adel Bibi. 2023.
\newblock \href {https://doi.org/10.52202/075280-1608} {Language model tokenizers introduce unfairness between languages}.
\newblock In \emph{Advances in Neural Information Processing Systems}, volume~36, pages 36963--36990. Curran Associates, Inc.

\bibitem[{Pires et~al.(2019)Pires, Schlinger, and Garrette}]{pires-etal-2019-multilingual}
Telmo Pires, Eva Schlinger, and Dan Garrette. 2019.
\newblock \href {https://doi.org/10.18653/v1/P19-1493} {How multilingual is multilingual {BERT}?}
\newblock In \emph{Proceedings of the 57th Annual Meeting of the Association for Computational Linguistics}, pages 4996--5001, Florence, Italy. Association for Computational Linguistics.

\bibitem[{Raffel et~al.(2020)Raffel, Shazeer, Roberts, Lee, Narang, Matena, Zhou, Li, and Liu}]{JMLR:v21:20-074}
Colin Raffel, Noam Shazeer, Adam Roberts, Katherine Lee, Sharan Narang, Michael Matena, Yanqi Zhou, Wei Li, and Peter~J. Liu. 2020.
\newblock \href {http://jmlr.org/papers/v21/20-074.html} {Exploring the limits of transfer learning with a unified text-to-text transformer}.
\newblock \emph{Journal of Machine Learning Research}, 21(140):1--67.

\bibitem[{Rahimi et~al.(2019)Rahimi, Li, and Cohn}]{rahimi-etal-2019-massively}
Afshin Rahimi, Yuan Li, and Trevor Cohn. 2019.
\newblock \href {https://doi.org/10.18653/v1/P19-1015} {Massively multilingual transfer for {NER}}.
\newblock In \emph{Proceedings of the 57th Annual Meeting of the Association for Computational Linguistics}, pages 151--164, Florence, Italy. Association for Computational Linguistics.

\bibitem[{Remy et~al.(2024)Remy, Delobelle, Avetisyan, Khabibullina, de~Lhoneux, and Demeester}]{remy2024transtokenization}
Fran{\c{c}}ois Remy, Pieter Delobelle, Hayastan Avetisyan, Alfiya Khabibullina, Miryam de~Lhoneux, and Thomas Demeester. 2024.
\newblock \href {https://openreview.net/forum?id=sBxvoDhvao} {Trans-tokenization and cross-lingual vocabulary transfers: Language adaptation of {LLM}s for low-resource {NLP}}.
\newblock In \emph{First Conference on Language Modeling}.

\bibitem[{Rust et~al.(2021)Rust, Pfeiffer, Vuli{\'c}, Ruder, and Gurevych}]{rust-etal-2021-good}
Phillip Rust, Jonas Pfeiffer, Ivan Vuli{\'c}, Sebastian Ruder, and Iryna Gurevych. 2021.
\newblock \href {https://doi.org/10.18653/v1/2021.acl-long.243} {How good is your tokenizer? on the monolingual performance of multilingual language models}.
\newblock In \emph{Proceedings of the 59th Annual Meeting of the Association for Computational Linguistics and the 11th International Joint Conference on Natural Language Processing (Volume 1: Long Papers)}, pages 3118--3135, Online. Association for Computational Linguistics.

\bibitem[{Sai~B et~al.(2023)Sai~B, Dixit, Nagarajan, Kunchukuttan, Kumar, Khapra, and Dabre}]{sai-b-etal-2023-indicmt}
Ananya Sai~B, Tanay Dixit, Vignesh Nagarajan, Anoop Kunchukuttan, Pratyush Kumar, Mitesh~M. Khapra, and Raj Dabre. 2023.
\newblock \href {https://doi.org/10.18653/v1/2023.acl-long.795} {{I}ndic{MT} eval: A dataset to meta-evaluate machine translation metrics for {I}ndian languages}.
\newblock In \emph{Proceedings of the 61st Annual Meeting of the Association for Computational Linguistics (Volume 1: Long Papers)}, pages 14210--14228, Toronto, Canada. Association for Computational Linguistics.

\bibitem[{Sennrich et~al.(2016)Sennrich, Haddow, and Birch}]{sennrich-etal-2016-neural}
Rico Sennrich, Barry Haddow, and Alexandra Birch. 2016.
\newblock \href {https://doi.org/10.18653/v1/P16-1162} {Neural machine translation of rare words with subword units}.
\newblock In \emph{Proceedings of the 54th Annual Meeting of the Association for Computational Linguistics (Volume 1: Long Papers)}, pages 1715--1725, Berlin, Germany. Association for Computational Linguistics.

\bibitem[{Tenney et~al.(2019)Tenney, Das, and Pavlick}]{tenney-etal-2019-bert}
Ian Tenney, Dipanjan Das, and Ellie Pavlick. 2019.
\newblock \href {https://doi.org/10.18653/v1/P19-1452} {{BERT} rediscovers the classical {NLP} pipeline}.
\newblock In \emph{Proceedings of the 57th Annual Meeting of the Association for Computational Linguistics}, pages 4593--4601, Florence, Italy. Association for Computational Linguistics.

\bibitem[{Wolf et~al.(2020)Wolf, Debut, Sanh, Chaumond, Delangue, Moi, Cistac, Rault, Louf, Funtowicz, Davison, Shleifer, von Platen, Ma, Jernite, Plu, Xu, Le~Scao, Gugger, Drame, Lhoest, and Rush}]{wolf-etal-2020-transformers}
Thomas Wolf, Lysandre Debut, Victor Sanh, Julien Chaumond, Clement Delangue, Anthony Moi, Pierric Cistac, Tim Rault, Remi Louf, Morgan Funtowicz, Joe Davison, Sam Shleifer, Patrick von Platen, Clara Ma, Yacine Jernite, Julien Plu, Canwen Xu, Teven Le~Scao, Sylvain Gugger, and 3 others. 2020.
\newblock \href {https://doi.org/10.18653/v1/2020.emnlp-demos.6} {Transformers: State-of-the-art natural language processing}.
\newblock In \emph{Proceedings of the 2020 Conference on Empirical Methods in Natural Language Processing: System Demonstrations}, pages 38--45, Online. Association for Computational Linguistics.

\bibitem[{Wu and Dredze(2020)}]{wu-dredze-2020-languages}
Shijie Wu and Mark Dredze. 2020.
\newblock \href {https://doi.org/10.18653/v1/2020.repl4nlp-1.16} {Are all languages created equal in multilingual {BERT}?}
\newblock In \emph{Proceedings of the 5th Workshop on Representation Learning for NLP}, pages 120--130, Online. Association for Computational Linguistics.

\bibitem[{Xue et~al.(2022)Xue, Barua, Constant, Al-Rfou, Narang, Kale, Roberts, and Raffel}]{xue-etal-2022-byt5}
Linting Xue, Aditya Barua, Noah Constant, Rami Al-Rfou, Sharan Narang, Mihir Kale, Adam Roberts, and Colin Raffel. 2022.
\newblock \href {https://doi.org/10.1162/tacl_a_00461} {{B}y{T}5: Towards a token-free future with pre-trained byte-to-byte models}.
\newblock \emph{Transactions of the Association for Computational Linguistics}, 10:291--306.

\bibitem[{Xue et~al.(2021)Xue, Constant, Roberts, Kale, Al-Rfou, Siddhant, Barua, and Raffel}]{xue-etal-2021-mt5}
Linting Xue, Noah Constant, Adam Roberts, Mihir Kale, Rami Al-Rfou, Aditya Siddhant, Aditya Barua, and Colin Raffel. 2021.
\newblock \href {https://doi.org/10.18653/v1/2021.naacl-main.41} {m{T}5: A massively multilingual pre-trained text-to-text transformer}.
\newblock In \emph{Proceedings of the 2021 Conference of the North American Chapter of the Association for Computational Linguistics: Human Language Technologies}, pages 483--498, Online. Association for Computational Linguistics.

\bibitem[{Yamaguchi et~al.(2025)Yamaguchi, Morishita, Villavicencio, and Aletras}]{yamaguchi2025adapting}
Atsuki Yamaguchi, Terufumi Morishita, Aline Villavicencio, and Nikolaos Aletras. 2025.
\newblock \href {https://openreview.net/forum?id=6IdoIKowfe} {Adapting chat language models using only target unlabeled language data}.
\newblock \emph{Transactions on Machine Learning Research}.

\bibitem[{Yamaguchi et~al.(2024)Yamaguchi, Villavicencio, and Aletras}]{yamaguchi-etal-2024-empirical}
Atsuki Yamaguchi, Aline Villavicencio, and Nikolaos Aletras. 2024.
\newblock \href {https://doi.org/10.18653/v1/2024.findings-emnlp.396} {An empirical study on cross-lingual vocabulary adaptation for efficient language model inference}.
\newblock In \emph{Findings of the Association for Computational Linguistics: EMNLP 2024}, pages 6760--6785, Miami, Florida, USA. Association for Computational Linguistics.

\bibitem[{Yamaguchi et~al.(2026)Yamaguchi, Villavicencio, and Aletras}]{yamaguchi-etal-2026-effectively}
Atsuki Yamaguchi, Aline Villavicencio, and Nikolaos Aletras. 2026.
\newblock \href {https://doi.org/10.1162/coli.a.581} {How can we effectively expand the vocabulary of {LLM}s with 0.01{GB} of target language text?}
\newblock \emph{Computational Linguistics}, 52(1):295--330.

\bibitem[{Yang et~al.(2025)Yang, Li, Yang, Zhang, Hui, Zheng, Yu, Gao, Huang, Lv, Zheng, Liu, Zhou, Huang, Hu, Ge, Wei, Lin, Tang, Yang, Tu, Zhang, Yang, Yang, Zhou, Zhou, Lin, Dang, Bao, Yang, Yu, Deng, Li, Xue, Li, Zhang, Wang, Zhu, Men, Gao, Liu, Luo, Li, Tang, Yin, Ren, Wang, Zhang, Ren, Fan, Su, Zhang, Zhang, Wan, Liu, Wang, Cui, Zhang, Zhou, and Qiu}]{yang2025qwen3technicalreport}
An~Yang, Anfeng Li, Baosong Yang, Beichen Zhang, Binyuan Hui, Bo~Zheng, Bowen Yu, Chang Gao, Chengen Huang, Chenxu Lv, Chujie Zheng, Dayiheng Liu, Fan Zhou, Fei Huang, Feng Hu, Hao Ge, Haoran Wei, Huan Lin, Jialong Tang, and 41 others. 2025.
\newblock \href {https://arxiv.org/abs/2505.09388} {Qwen3 technical report}.
\newblock \emph{Preprint}, arXiv:2505.09388.

\bibitem[{YU et~al.(2023)YU, Simig, Flaherty, Aghajanyan, Zettlemoyer, and Lewis}]{NEURIPS2023_f8f78f80}
LILI YU, Daniel Simig, Colin Flaherty, Armen Aghajanyan, Luke Zettlemoyer, and Mike Lewis. 2023.
\newblock \href {https://doi.org/10.52202/075280-3447} {Megabyte: Predicting million-byte sequences with multiscale transformers}.
\newblock In \emph{Advances in Neural Information Processing Systems}, volume~36, pages 78808--78823. Curran Associates, Inc.

\end{thebibliography}

\appendix

\section{Appendix}
\label{sec:appendix}

\subsection{Language complexity} \label{sec:lang_complexity}
The selected languages span a broad range of morphological complexity and script variation. Hindi and Urdu are closely related languages that are largely mutually intelligible in speech, differing primarily in script (Devanagari vs.\ Nastaliq) and formal vocabulary. Bhojpuri and Magahi are closely related to Hindi, sharing similar morphological structure and the Devanagari script, but lack standardized orthography and large-scale NLP resources. Marathi exhibits richer nominal and verbal inflection than Hindi, including a larger inventory of case markers. Sanskrit is the most morphologically complex language in our study: a classical language with highly synthetic morphology, extensive sandhi phenomena (phonological fusion at word boundaries), and relatively free word order, making it particularly challenging for both tokenization and byte-level boundary detection. Urdu presents an additional challenge due to its Nastaliq script, which differs substantially from the Devanagari script used in the pretraining data.

\subsection{Language and Dataset}

We evaluate across five Indic languages spanning two scripts and varying levels of morphological complexity, resource availability, and typological proximity to Hindi. Bhojpuri and Magahi are extremely low-resource Indo-Aryan languages closely related to Hindi and written in Devanagari, making them suitable for evaluating cross-lingual transfer under severe tokenizer overfragmentation. Marathi also uses Devanagari but exhibits richer inflectional morphology and greater lexical divergence from Hindi. Sanskrit is highly morphologically rich, with extensive inflection and compounding, leading to particularly severe subword fragmentation. Urdu differs both linguistically and script-wise, using the Perso-Arabic Nastaliq script rather than Devanagari, which poses a challenging setting for cross-script transfer.

For POS tagging, we use Universal Dependencies (UD) treebanks~\citep{nivre-etal-2020-universal}. Hindi (\textsc{hi-hdtb}) serves as the source language and contains 13,306 training sentences and 1,684 test sentences (281,057 and 35,430 words, respectively). The remaining languages are substantially smaller: Bhojpuri (\textsc{bho-bhtb}, 357 sentences, 6,665 words), Marathi (\textsc{mr-ufal}, 373 sentences, 3,253 words), Magahi (\textsc{mag-mgtb}, 550 sentences, 7,702 words), Sanskrit (\textsc{sa-ufal}, 230 sentences, 2,102 words), and Urdu (\textsc{ur-udtb}, 535 sentences, 14,806 words). These dataset sizes reflect the extremely limited supervision available for many Indic languages.

For NER, we use WikiANN~\citep{rahimi-etal-2019-massively}. Hindi contains 5,000 training sentences, while Urdu and Marathi each contain 1,000 test sentences, and Sanskrit contains only 100 test sentences. For sentiment analysis, we use IndicSentiment~\citep{doddapaneni-etal-2023-towards} with 1,000 examples each for Hindi, Marathi, and Urdu using binary positive/negative classification.

Together, these datasets provide evaluation across morphologically grounded sequence-labeling tasks (POS and NER) and semantic classification (sentiment analysis), under both cross-lingual and cross-script transfer settings.

\subsection{Training Details}
\label{sec:training_details}
\begin{table}[h]
\centering
\scriptsize
\setlength{\tabcolsep}{2pt}
\resizebox{\linewidth}{!}{
\begin{tabular}{llc}
\toprule
\textbf{Setting} & \textbf{Hyperparameter} & \textbf{Value} \\
\midrule
\multirow{13}{*}{H-Net pretrain}
  & Optimiser              & AdamW \\
  & Learning rate          & $5\times10^{-5}$ \\
  & Weight decay           & 0.01 \\
  & Max grad norm          & 0.5 \\
  & LR schedule            & Cosine + 5\% warmup \\
  & Batch size             & 8 sequences \\
  & Sequence length        & 2,048 bytes \\
  & Precision              & bfloat16 \\
  & $\lambda_{\text{ratio}}$      & 0.8 (follow~\citet{hwang2026dynamic}) \\
  & $\lambda_{\text{align}}$      & 0.3 \\
  & $\lambda_{\text{POS}}$        & 0.3 \\
  & POS frequency          & every 10 AR steps \\
  & Target rate $\rho^{*}$ & 0.056 \\
  & Mamba d\_state         & 16 \\
  & Mamba d\_conv          & 4 \\
  & Mamba expand (Qwen)    & 2 \\
  & Mamba expand (Gemma)   & 1 \\
\midrule
\multirow{5}{*}{CPT baseline}
  & Optimiser              & AdamW \\
  & Learning rate          & $5\times10^{-5}$ \\
  & Weight decay           & 0.01 \\
  & LR schedule            & Cosine + 5\% warmup \\
  & Unfrozen layers        & first \& last \\
\midrule
\multirow{5}{*}{Probe (POS/NER)}
  & Optimiser              & AdamW \\
  & Learning rate          & $2\times10^{-4}$ \\
  & Weight decay           & 0.01 \\
  & Batch size             & 32 \\
  & Epochs                 & 10 \\
\midrule
\multirow{5}{*}{Probe (Sentiment)}
  & Optimiser              & AdamW \\
  & Learning rate          & $2\times10^{-4}$ \\
  & Weight decay           & 0.01 \\
  & Batch size             & 16 \\
  & Epochs                 & 10 \\
\bottomrule
\end{tabular}
}
\caption{Hyperparameters for H-Net pretraining, CPT
baseline, and downstream probe training.}
\label{tab:hyperparams}
\end{table}
We pretrain H-Net using AdamW ($\text{lr} = 5\times10^{-5}$, weight\_decay $= 0.01$, max\_grad\_norm $= 0.5$) with cosine learning rate decay and 5\% linear warmup. We use a batch size of 8 sequences of 2,048 bytes (2 sequences $\times$ 4 gradient accumulation steps) in bfloat16 precision. The local encoder and decoder each use a single Mamba layer (d\_state $= 16$, d\_conv $= 4$, expand $= 2$ for Qwen; expand $= 1$ for Gemma). For downstream evaluation, we train a two-layer MLP probe (d\_model $\to$ 256 $\to$ n\_tags, GELU, LayerNorm) using AdamW ($\text{lr} = 2\times10^{-4}$, weight\_decay $= 0.01$, cosine decay, 5\% warmup) for 10 epochs (POS, NER) or 10 epochs (sentiment). All remaining hyperparameters are listed in Table~\ref{tab:hyperparams}. 

All H-Net byte-level components are pretrained on a single NVIDIA A100 GPU, taking approximately 26 hours per backbone. Downstream probe training is performed once on Hindi and evaluated directly on all target languages without retraining, taking approximately 10--30 minutes.

\subsection{In-Language Hindi Evaluation}
\label{app:hindi}

Table~\ref{tab:hindi_eval} reports in-language Hindi evaluation on the held-out UD Hindi test set. For NER, H-Net outperforms the subword baseline across all models ($+$5.3, $+$9.4, $+$3.2~pp), confirming that byte-level chunk representations benefit entity boundary detection even at lower fragmentation rates (2.61~tok/word). For POS tagging, H-Net underperforms on Hindi for most models. Hindi is the adaptation language with relatively low fragmentation, so the subword baseline has a full advantage in the supervised setting. The zero-shot setting reverses this pattern, with H-Net winning across all five target languages where  fragmentation of the standard tokenizer is higher (3.02--5.72~tok/word). For sentiment, H-Net underperforms consistently relative to target-language results, confirming that semantic tasks benefit more from the training data scale rather than effectively processing their morphological structure.

\begin{table}[h]
\centering
\small
\setlength{\tabcolsep}{4pt}
\begin{tabular}{llcc}
\toprule
\textbf{Task} & \textbf{Model} & \textbf{Sub} & \textbf{H-Net} \\
\midrule
\multirow{3}{*}{POS}
  & Qwen3-1.7B & \textbf{65.3} & 59.5 \\
  & Qwen3-4B   & 71.6 & \textbf{72.1} \\
  & Gemma3-4B  & \textbf{94.3} & 87.4 \\
\midrule
\multirow{3}{*}{NER}
  & Qwen3-1.7B & 24.6 & \textbf{29.9} \\
  & Qwen3-4B   & 25.2 & \textbf{34.6} \\
  & Gemma3-4B  & 52.2 & \textbf{55.4} \\
\midrule
\multirow{3}{*}{Sentiment}
  & Qwen3-1.7B & \textbf{88.5} & 71.5 \\
  & Qwen3-4B   & \textbf{89.5} & 84.5 \\
  & Gemma3-4B  & \textbf{95.0} & 88.5 \\
\bottomrule
\end{tabular}
\caption{In-language Hindi evaluation on the held-out UD Hindi test set. Bold indicates the best result per row.}
\label{tab:hindi_eval}
\end{table}

\section{Use of Generative AI Tools}
The authors acknowledge the use of LLMs in preparing this work. In particular, GPT-4o was used to improve the grammar, clarity, and organization of the manuscript, as well as to assist with code debugging. All technical content, experimental design, implementation decisions, and analysis are implemented and verified by the authors.

\end{document}